\documentclass{article}
\usepackage{ijcai26}

\usepackage{times}
\usepackage{soul}
\usepackage{url}
\usepackage[hidelinks]{hyperref}
\usepackage[utf8]{inputenc}
\usepackage[small]{caption}
\usepackage{graphicx}
\usepackage{amsmath}
\usepackage{amsthm}
\usepackage{booktabs}
\usepackage{algorithm}
\usepackage{algorithmic}
\usepackage[switch]{lineno}
\usepackage{amssymb}
\usepackage{enumitem}
\usepackage{booktabs}  
\usepackage{xcolor}    
\usepackage{graphicx}  
\usepackage{multirow}  
\usepackage{pifont}
\usepackage{amssymb}
\usepackage{cuted}
\usepackage{caption}

\definecolor{DarkGreen}{rgb}{0, 0.4, 0}

\title{Rethinking Multimodal Few-Shot 3D Point Cloud Segmentation: From Fused Refinement to Decoupled Arbitration}

\author{
    Wentao Bian, Fenglei Xu\thanks{Corresponding author} \\
    {\large \normalfont Suzhou University of Science and Technology, Suzhou, China} \\
    {\large \normalfont bwt1819548635@gmail.com, xufl@mail.usts.edu.cn}
}

\begin{document}

\maketitle

\begin{abstract}
    In this paper, we revisit multimodal few-shot 3D point cloud semantic segmentation (FS-PCS), identifying a conflict in "Fuse-then-Refine" paradigms: the "Plasticity-Stability Dilemma." In addition, CLIP's inter-class confusion can result in semantic blindness. To address these issues, we present the \textbf{D}ecoupled-experts \textbf{A}rbitration \textbf{F}ew-\textbf{S}hot \textbf{S}egNet (\textbf{DA-FSS}), a model that effectively distinguishes between semantic and geometric paths and mutually regularizes their gradients to achieve better generalization. DA-FSS employs the same backbone and pre-trained text encoder as MM-FSS to generate text embeddings, which can increase free modalities' utilization rate and better leverage each modality's information space. To achieve this, we propose a Parallel Expert Refinement module to generate each modal correlation. We also propose a \textbf{S}tacked \textbf{A}rbitration \textbf{M}odule (\textbf{SAM}) to perform convolutional fusion and arbitrate correlations for each modality pathway. The Parallel Experts decouple two paths: a Geometric Expert maintains plasticity, and a Semantic Expert ensures stability. They are coordinated via a \textbf{D}ecoupled \textbf{A}lignment \textbf{M}odule (\textbf{DAM}) that transfers knowledge without propagating confusion. Experiments on popular datasets (S3DIS, ScanNet) demonstrate the superiority of DA-FSS over MM-FSS. Meanwhile, geometric boundaries, completeness, and texture differentiation are all superior to the baseline. The code is available at: \url{https://github.com/MoWenQAQ/DA-FSS/}.
\end{abstract}

\section{Introduction} 
3D point cloud analysis is following the development of \textbf{Label-efficient learning}~\cite{xiao2024survey,chen2023clip2scene} and the advent of Large Vision-Language Models (VLMs) and 3D-LLMs~\cite{hong20243dllm} has triggered a revolution in 3D scene understanding ~\cite{xu2024pointllm,Xue_2024_CVPR,guo2023point,hong20243dllm,liang2024pointmamba}. Through establishing a connection between point clouds and human language, some methods have broken the recognition bottleneck imposed by the closed label sets of traditional supervised learning~\cite{qi2017pointnet,wang2019dynamic,wu2024ptv3}. Meanwhile, the paradigm shift towards leveraging foundation models is exemplified by Point-Bind~\cite{guo2023point} and ULIP-2~\cite{Xue_2024_CVPR,xue2023ulip} which are two examples of aggregated frameworks that have successfully created a \textbf{shared embedding space} for 3D and human language. For instance, OpenScene~\cite{peng2023openscene}, OV3D~\cite{jiang2024ov3d}, RegionPLC~\cite{yang2024regionplc}, UniM-OV3D~\cite{he2024unim}, Open3DIS~\cite{nguyen2024open3dis}, and PLA~\cite{ding2023pla} all demonstrate solid zero-shot generalization by building upon these semantic priors. Despite these advances, Few-Shot Learning (FSL) ~\cite{snell2017prototypical,zhao2021few,hong2022cost} is still a critical challenge due to annotated scarce data.

In the realm of MM-FS-PCS, our key insight is that \textbf{MM-FSS}~\cite{an2024multimodality} is the pioneering \textbf{SOTA} that validated the utility of VLM priors. However, we’ve found that MM-FSS suffers from a \textit{Plasticity-Stability Dilemma} which creates a \textbf{Gradient Domination}. This is caused by CLIP’s category confusion ~\cite{li2025logits,zhu2023not,tu2023closer,qiu2024shape}, which leads to \textit{semantic hallucinations} in geometrically ambiguous regions (e.g., confusing a white wall with a picture).

Specifically, CLIP (Contrastive Language-Image Pre-training) produces semantic features that have been trained at a large scale. Its vector norms are typically large and stable. MM-FSS chooses to freeze the parameters of CLIP (IF Head) to preserve semantic features, leveraging the VLM, aligning with the trend of parameter-efficient fine-tuning~\cite{tang2024point,zhang2022tip}. This leads to the problem that the model will discover that "I can achieve a high score simply by copying CLIP's semantic outputs." Thereby causing the 3D encoder to learn from 2D features in a simplistic manner, resulting in feature representations with relatively smaller norms, with semantic features dominating gradient flow.

Motivated by these important observations, a pertinent question arises: How can we limit gradient conflict and prevent category confusion? In this paper, the idea comes from the philosophy of Logits DeConfusion~\cite{li2025logits}, which physically separates confusion noise through residual learning. So we suggest a new paradigm \textbf{``Decoupled Arbitration''} to resolve this problem. We believe that geometric adaptation and semantic preservation need pathways that are completely separate from each other. Interaction should be limited to two laws: \textbf{Soft Regularization} which guide the geometric expert without messing up its input and \textbf{Late Arbitration} which only combines decisions after the decoupled decisions have been independently refined. Specifically, we introduce a novel model, \textbf{Decoupled-experts Arbitration Few-Shot SegNet (DA-FSS)}, to leverage multimodal information more efficiently. Concretely, we develop an \textbf{Adaptive Geometric Expert} and a \textbf{Static Semantic Expert} to form the Decoupled-experts. The following \textbf{S}tacked \textbf{A}rbitration \textbf{M}odule (\textbf{SAM}) performs convolutional fusion of the independently refined features from both pathways, building upon utilizing semantic guidance from textual information. Additionally, we propose \textbf{D}ecoupled \textbf{A}lignment \textbf{M}odule (\textbf{DAM}) enforcing the geometric expert to align with the semantic expert at the representation level so as to prevent the decoupling from causing each path to deviate excessively.

We systematically compare our DA-FSS against existing methods AttMPTI~\cite{zhao2021few}, QGE~\cite{ning2023boosting}, COSeg~\cite{an2024rethinking}, and MM-FSS~\cite{an2024multimodality} on S3DIS~\cite{armeni20163d} and ScanNet~\cite{dai2017scannet} datasets ~\ref{sec:comparison}, suggesting the superiority of DA-FSS across various settings.

The contributions of our work are threefold: (i) We design a novel model, DA-FSS, to more effectively exploit information from different modalities. To the best of our knowledge, this is the first work to explore decoupling in MM-FSS. (ii) Learning from the baseline’s novel cost-free multimodal FS-PCS design philosophy, our DA-FSS efficiently reuses the pre-trained alignment already completed by the baseline to achieve optimization without any extra parameter overhead. (iii) Extensive experimental results demonstrate the value of our proposed framework and its consistent efficacy across different few-shot settings. This work will provide a new perspective for other multimodal fusion approaches.

\section{Related Work} 
\label{sec:related_work} 
\subsection{Few-Shot 3D Point Cloud Segmentation} 
Although fully-supervised methods have achieved promising results in 3D point cloud segmentation~\cite{qi2017pointnet,qi2017pointnet2,wang2019dynamic,lai2022stratified,wu2024ptv3,yu2022point}, the expensive annotation costs have shifted research attention toward few-shot learning. As a pioneer, AttMPTI~\cite{zhao2021few} proposed a transductive label propagation method. Later works mainly concentrate on narrowing the semantic gap~\cite{zhang2023few,ning2023boosting,ning2023boosting,xiong2024aggregation} and improving feature representation~\cite{an2024rethinking}. Specifically, AttMPTI~\cite{zhao2021few} and COSeg~\cite{an2024rethinking} optimize prototype correlations to reduce intra-class variance. In contrast, SCAT~\cite{zhang2023few} and QGE~\cite{ning2023boosting} apply advanced cross-attention mechanisms for fine-grained alignment between support and query features. However, recent observations indicate that these methods rely mainly on geometric features, which limits their generalization in complex open-world scenes (e.g., distinguishing objects with similar geometry). 

\subsection{Multimodal 3D Point Cloud Segmentation} 
With the rapid development of unimodal baselines, researchers are increasingly exploring multimodality for further performance gains. Language is currently the most popular auxiliary modality. Unified frameworks~\cite{guo2023point,Xue_2024_CVPR,xu2024pointllm,zhou2023uni3d} align 3D point clouds with image-text spaces to boost performance . Furthermore, open-vocabulary methods (e.g., OV3D~\cite{jiang2024ov3d}, RegionPLC~\cite{yang2024regionplc}, Open3DIS~\cite{nguyen2024open3dis}, UniM-OV3D~\cite{he2024unim}, PointTFA~\cite{wu2024pointtfa}) utilize VLM guidance to achieve open-set segmentation. Nevertheless, these approaches are tailored primarily for fully supervised or zero-shot settings. In the context of FS-PCS, existing works still predominantly use unimodal input. Recently, MM-FSS~\cite{an2024multimodality} and Generalized FS-PCS~\cite{an2025generalized} attempt to leverage multimodal information. Critically, these methods follow a ``Fused Refinement'' paradigm. We observe that the aggressive fusion in these works leads to \textit{Gradient Domination}. To the best of our knowledge, we are the first to introduce decoupled expert arbitration~\cite{li2025logits,zhang2022tip,zhao2022decoupled} into multimodal FS-PCS.

\begin{figure}[t]
    \centering
    \includegraphics[width=0.95\linewidth]{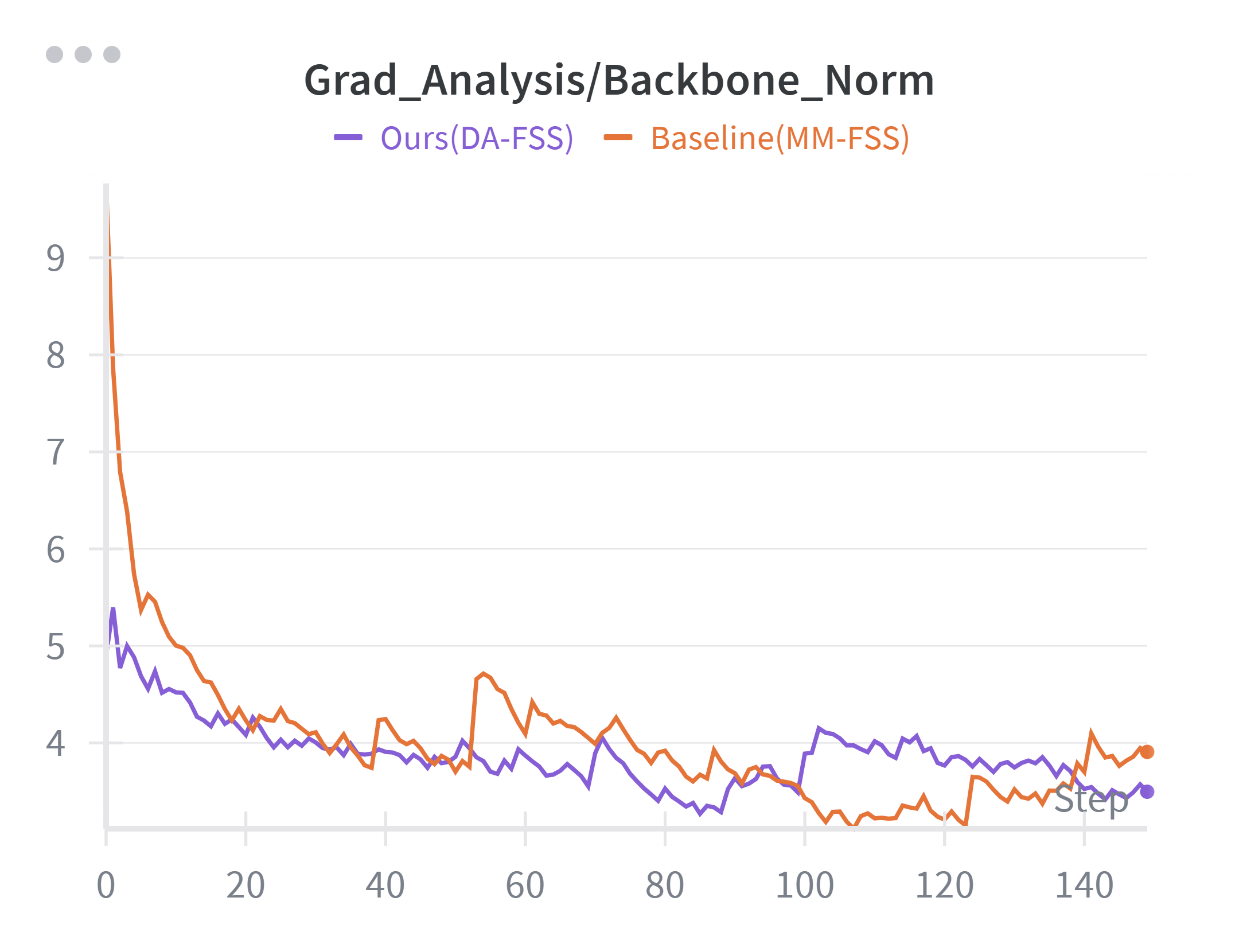} 
    \vspace{-2mm}
    \caption{The Baseline (Orange)’s gradient norm declines rapidly with high volatility, characterized by diminishing gradients due to semantic domination. In contrast, our DA-FSS (Purple)’s gradient norm maintains a relatively stable trend.}
    \label{fig:grad_analysis}
    \vspace{-4mm}
\end{figure}

\section{Methodology}
\label{sec:methodology}

\begin{figure*}[t]
    \centering
    \includegraphics[width=0.95\textwidth]{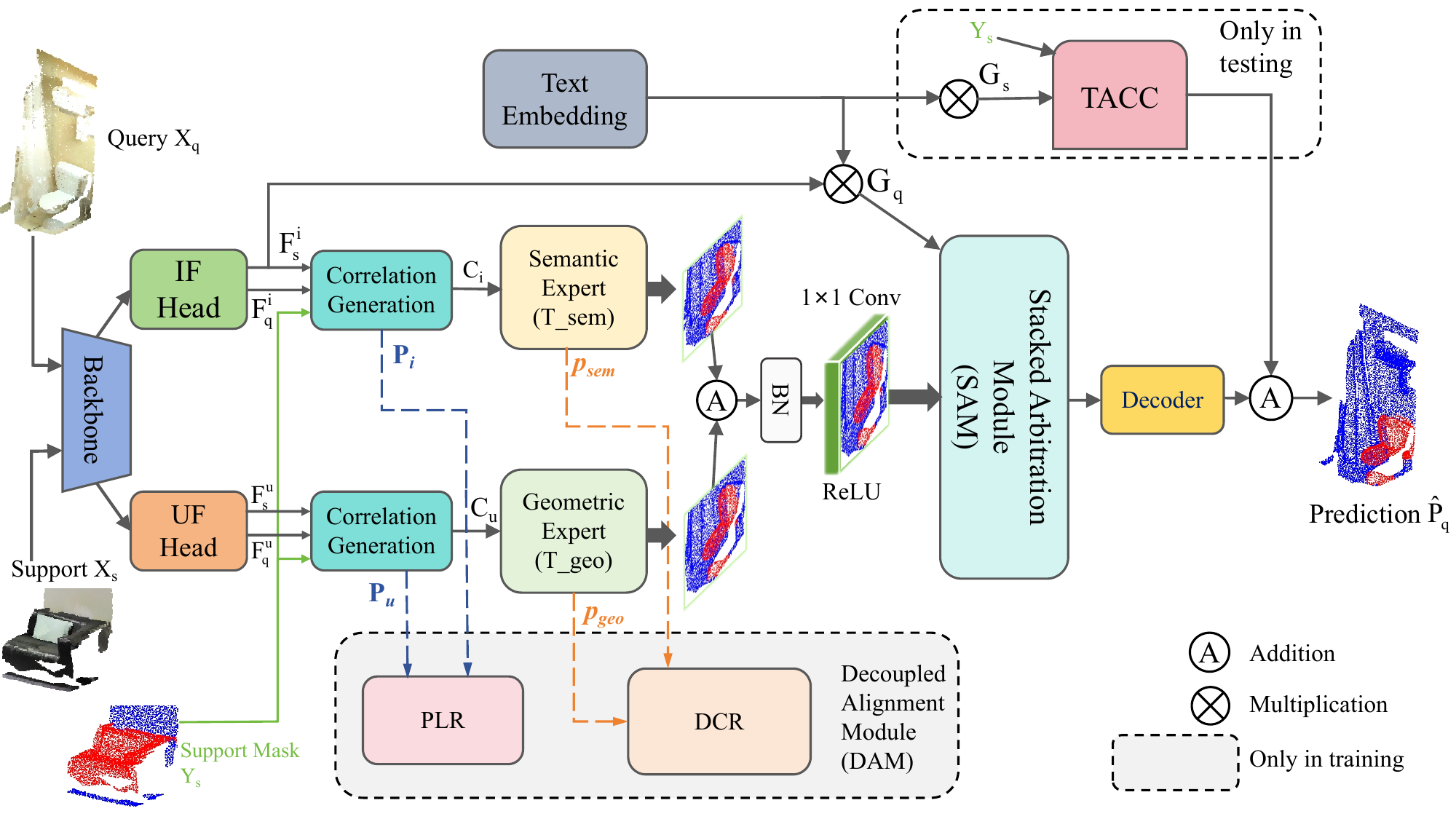}
    \vspace{-2mm}
    \caption{\textbf{Overall architecture of the proposed DA-FSS.} Given support and query point clouds, we first generate intermodal correlations $F_{s/q}^{i}$ from the IF head and unimodal correlations $F_{s/q}^{u}$ from the UF head. These correlations are then forwarded to the \textbf{Parallel Experts} (Geometric Expert $\mathcal{T}_{geo}$ and Semantic Expert $\mathcal{T}_{sem}$) to independently refine features, isolating plasticity from stability. Moreover, we use the \textbf{Decoupled Alignment Module (DAM)} to align correlations ($C_i, C_u$) via PLR and refined experts ($P_{sem}, P_{geo}$) via DCR \textbf{only in training}. Finally, we propose the \textbf{Stacked Arbitration Module (SAM)}, which synthesizes the final decision by leveraging boundary-injected guidance ($G_{base}$ and $G_q$) to effectively arbitrate between geometric and semantic pathways after a 1$\times$1 convolutional fusion. For clarity, we present the model under the 1-way 1-shot setting.}
    \label{fig:architecture}
    \vspace{-4mm}
\end{figure*}

\subsection{Problem Setup} 
\label{sec: Setup } 
\paragraph{Multimodal FS-PCS.} Following prior works~\cite{an2024rethinking,an2024multimodality}, we formulate this task as the popular episodic paradigm. Each $N$-way $K$-shot episode comprises a support set $\mathcal{S} = \{X_{s}^{n,k}, Y_{s}^{n,k}\}$ and a query set $\mathcal{Q} = \{X_{q}, Y_{q}\}$, where $X$ and $Y$ denote the point cloud and corresponding labels, respectively. The goal is to segment query samples $X_{q}$ into $N$ target classes and `background' by leveraging knowledge from $\mathcal{S}$. Following the multimodal setting, our DA-FSS introduces textual and 2D modalities.

\subsection{Motivation} 
\label{sec: Motivation} 
Current SOTA methods~\cite{an2024multimodality} typically adopt an early fusion strategy, obtaining a joint representation $H = \sigma(W \cdot [F_{geo} \oplus F_{sem}])$.
Although this is a simple and intuitive approach, we argue that this paradigm creates a fundamental optimization conflict : during meta-training, the gradient flow to the geometric encoder called \textbf{UF Head}~\cite{an2024multimodality} is always coupled with the semantic branch called \textbf{IF Head}~\cite{an2024multimodality}:
\begin{equation}
    \frac{\partial \mathcal{L}}{\partial F_{UF}} = \frac{\partial \mathcal{L}}{\partial H} \cdot \frac{\partial H}{\partial F_{UF}}
\end{equation}

We identify two issues in this interaction: (1) The static semantic features $F_{sem}$ (from frozen VLMs) typically exhibit significantly larger norms than the adaptive geometric features, numerically overwhelming the sparse geometric gradients. (2) Frozen VLMs suffer from inherent \textit{texture bias}~\cite{li2025logits,tu2023closer}. In geometrically ambiguous regions, the optimization shortcut provided by texture semantics suppresses the learning of structural details.

Consequently, the \textbf{UF Head} fails to adapt effectively, a phenomenon we term \textit{Plasticity Decay}: 
\begin{equation} 
    \left\| \nabla_{\theta_{UF}} \mathcal{L} \right\| \xrightarrow{training} \delta_{low} \ll \left\| \nabla_{\text{init}} \right\| \label{eq:conflict} 
\end{equation} 

We provide concrete evidence of this phenomenon in Figure~\ref{fig:grad_analysis}. By monitoring the training dynamics, we observe that the baseline's geometric gradients (Orange) are strictly suppressed by the semantic branch, effectively rendering the backbone ``lazy'' so that the \textbf{UF Head} will only obey the VLM’s guidance from the \textbf{IF Head} (\textbf{Frozen}) without learning to optimize itself. This observation motivates us to \textbf{physically decouple} the optimization pathways to restore the backbone's vitality.

\subsection{Method Overview}
\label{subsec:our_approach}

The overall architecture of the proposed DA-FSS is depicted in Figure~\ref{fig:architecture}.
Given support and query point clouds, we first generate two sets of dense correlations: intermodal correlations from the Intermodal Feature (IF) head and unimodal correlations from the Unimodal Feature (UF) head. Both intermodal and unimodal correlations are then forwarded to the \textbf{Parallel Expert Refinement} module to independently produce refined geometric and semantic features. Beyond independent refinement, we use the \textbf{D}ecoupled \textbf{A}lignment \textbf{M}odule (\textbf{DAM}) to bridge the modal gap. This module acts as a soft regularizer to align the experts without gradient interference. We then exploit useful boundary-injected guidance to synthesize the final decision in the \textbf{S}tacked \textbf{A}rbitration \textbf{M}odule (\textbf{SAM}).

Existing MultiModal FS-PCS approaches~\cite{an2024multimodality} typically have two training steps: a pretraining step for obtaining an effective feature extractor, and a meta-learning step towards few-shot segmentation tasks. Our method follows this two-step training paradigm. First, we pretrain the backbone and IF head using 3D point clouds and 2D images. To more objectively demonstrate our improvement results, we chose to directly use the same method as MM-FSS does, or even download officially released pre-trained weights directly. Second, we conduct meta-learning to train the model end-to-end while freezing the backbone and IF head to maintain stability. In the following, we elaborate on Parallel Experts, DAM, and SAM modules.

\subsection{Parallel Experts Refinement}
\label{subsec:parallel_experts}

Inspired by Mixture-of-Experts (MoE)~\cite{jacobs1991adaptive}, we propose the Parallel Experts module, as illustrated in Figure~\ref{fig:architecture} to preserve the unique characteristics of each modality. Following ~\cite{an2024multimodality}, we first merge correlations to capture support-query interactions. The \textit{Unimodal Correlation} $C^u \in \mathbb{R}^{N_q \times N_s}$ captures geometric similarity, while the \textit{Intermodal Correlation} $C^i \in \mathbb{R}^{N_q \times N_s}$ captures semantic affinity using VLM projections. This enhanced comprehension promotes the transfer of knowledge from the support set to the query point cloud, thereby refining segmentation outcomes.MM-FSS chose to sum $C^u$ and $C^i$, but we do not opt for this approach; conversely, we propose that:

\vspace{1mm}
\noindent\textbf{Adaptive Geometric Expert ($\mathcal{T}_{geo}$).}
This expert is fully learnable and responsible for \textit{Plasticity} by receiving $C^u$, which focuses on capturing task-specific structural variations. To this end, we employ a standard Transformer Layer (incorporating MHSA)  to capture the holistic feature tensor of $H_{geo}$ (192-dimensional):
\begin{equation}
    \begin{aligned}
        H_{geo} &= \mathcal{F}_{lin1}(C^u) \\
        R_{geo} &= \text{LN}\left(H_{geo} + \text{MHSA}(H_{geo}, H_{geo}, H_{geo})\right) 
    \end{aligned}
\end{equation}

\vspace{1mm}
\noindent\textbf{Static Semantic Expert ($\mathcal{T}_{sem}$).}
This expert is fully learnable and responsible for \textit{Stability} by receiving $C^i$, which leverages the frozen VLM priors. This is to provide a consistent semantic reference robust to few-shot noise. Although this expert’s functions differ, we have chosen an identical design but $H_{sem}$ (512-dimensional):
\begin{equation}
    \begin{aligned}
        H_{sem} &= \mathcal{F}_{lin2}(C^i) \\
        R_{sem} &= \text{LN}\left(H_{sem} + \text{MHSA}(H_{sem}, H_{sem}, H_{sem})\right)
    \end{aligned}
\end{equation}
Crucially, by isolating these pathways ($\mathcal{I}(R_{geo}; R_{sem}) \approx 0$ during refinement), we ensure the experts are optimized solely based on features unique to their respective responsibilities, thereby protecting UF head against semantic gradient domination.

\subsection{Decoupled Alignment Module (DAM)}
\label{subsec:dam}
While the Parallel Experts module effectively isolates interference between modalities, the consistency would largely disappear due to decoupling, which could lead to feature divergence. However, we cannot forcibly intervene to rigidly align the features of the two modules. Therefore, we propose the DAM, as illustrated in Fig.~\ref{fig:architecture}. 
DAM integrates structural consensus from the semantic expert to regularize the geometric expert. 
Additionally, since raw semantic priors contain fixed confusion patterns~\cite{li2025logits}, DAM employs a stop-gradient strategy to filter out texture noise, accounting for the reliability gap between adaptive and frozen features.

\paragraph{Prototype Loss Regularization (PLR).}
Given learnable geometric prototypes $P_u$ and frozen semantic prototypes $P_i$, since the semantic prototypes $P_i$ originate from stable VLM priors, they provide informative guidance on the true semantic manifold. 
Therefore, we first compute the alignment distance between them to generate structural constraints $\mathcal{L}_{PLR}$. Specifically, we project $P_u$ (192-dimensional) to the dimension of $P_i$ (512-dimensional) via a linear layer $F_{proj}$ and minimize:
\begin{equation}
   \mathcal{L}_{PLR} = \frac{1}{|B_S|} \sum_{(P_u, P_i) \in B_S} \left\| F_{proj}(P_u) - \text{sg}(P_i) \right\|_2^2,
   \label{eq:plr}
\end{equation}
where $\text{sg}(\cdot)$ denotes the stop-gradient operator. Note that this effectively uses the semantic prototype as a "Teacher" anchor, ensuring the geometric encoder learns the semantic structure without inheriting confusion gradients.

\paragraph{Decoupled Consistency Regularization (DCR).}
Next, to enforce consistency at the decision boundary, we minimize the divergence between the probability distributions of the two experts ($p_{geo}, p_{sem}$). 
Since confusion patterns are often manifested in the logits distribution, the regularization term is computed as follows:
\begin{equation}
\begin{split}
\mathcal{L}_{DCR} = & \frac{1}{2} D_{KL}(p_{geo} \parallel \text{sg}(p_{sem})) \\
& + \frac{1}{2} D_{KL}(p_{sem} \parallel \text{sg}(p_{geo})).
\end{split}
\label{eq:dcr}
\end{equation}
Through this interaction, $\mathcal{T}_{sem}$ can also learn the specialized adaptation of the other branch, rather than solely optimizing itself. This is akin to how a "teacher" typically influences a student, yet the "student" can also subtly impact their own "teacher."

This DAM module fully leverages the stability of the semantic expert to guide the plasticity of the geometric expert, helping to maintain feature coherence. Note that it uses stop-gradient operators to prevent the propagation of confusion noise, improving the effective collaboration of decoupled experts.

\begin{table*}[t]
  \centering

  \resizebox{\textwidth}{!}{%
  \begin{tabular}{l ccc ccc ccc ccc}
    \toprule
    \multirow{2}{*}{Method} & \multicolumn{3}{c}{1-way 1-shot} & \multicolumn{3}{c}{1-way 5-shot} & \multicolumn{3}{c}{2-way 1-shot} & \multicolumn{3}{c}{2-way 5-shot} \\
    \cmidrule(lr){2-4} \cmidrule(lr){5-7} \cmidrule(lr){8-10} \cmidrule(lr){11-13}
    & $S^0$ & $S^1$ & Mean & $S^0$ & $S^1$ & Mean & $S^0$ & $S^1$ & Mean & $S^0$ & $S^1$ & Mean \\
    \midrule
    \multicolumn{13}{l}{\textit{Official Results (from Literature)}} \\
    AttMPTI~\cite{zhao2021few} & 36.32 & 38.36 & 37.34 & 46.71 & 42.70 & 44.71 & 31.09 & 29.62 & 30.36 & 39.53 & 32.62 & 36.08 \\
    QGE~\cite{ning2023boosting} & 41.69 & 39.09 & 40.39 & 50.59 & 46.41 & 48.50 & 33.45 & 30.95 & 32.20 & 40.53 & 36.13 & 38.33 \\
    COSeg~\cite{an2024rethinking} & 47.17 & 48.37 & 47.77 & 50.93 & 49.88 & 50.41 & 37.15 & 38.99 & 38.07 & 42.73 & 40.25 & 41.49 \\
    \textcolor{gray}{MM-FSS~\cite{an2024multimodality}} & \textcolor{gray}{49.84} & \textcolor{gray}{54.33} & \textcolor{gray}{52.09} & \textcolor{gray}{51.95} & \textcolor{gray}{56.46} & \textcolor{gray}{54.21} & \textcolor{gray}{41.98} & \textcolor{gray}{46.61} & \textcolor{gray}{44.30} & \textcolor{gray}{46.02} & \textcolor{gray}{54.29} & \textcolor{gray}{50.16} \\
    \midrule
    \multicolumn{13}{l}{\textit{Reproduced Environment (Strict Control)}} \\
    MM-FSS$^{\dagger}$ (Official Weights) & 48.50 & 54.15 & 51.33 & 51.76 & 56.43 & 54.10 & 41.70 & 47.00 & 44.35 & 45.67 & 54.82 & 50.25 \\ 
    \textbf{Ours (DA-FSS)} 
    & \textbf{49.14} & \textbf{55.93} & \textbf{52.54}$_{\textcolor{DarkGreen}{\uparrow\textbf{1.21}}}$ 
    & \textbf{51.90} & \textbf{60.45} & \textbf{56.18}$_{\textcolor{DarkGreen}{\uparrow\textbf{2.08}}}$ 
    & \textbf{42.07} & \textbf{47.16} & \textbf{44.62}$_{\textcolor{DarkGreen}{\uparrow\textbf{0.27}}}$ 
    & \textbf{45.78} & \textbf{56.30} & \textbf{51.04}$_{\textcolor{DarkGreen}{\uparrow\textbf{0.79}}}$ \\
    \bottomrule
  \end{tabular}
  }
  \caption{\textbf{Quantitative results on S3DIS.} \textit{Top:} Official results reported in literature. \textit{Bottom:} Comparison under our strictly controlled evaluation protocol using official pre-trained weights for the baseline (MM-FSS$^{\dagger}$). Our method (DA-FSS) consistently outperforms the baseline across most splits.}
    \label{tab:sota_s3dis}
\end{table*}

\begin{table*}[t]
  \centering

  \resizebox{\textwidth}{!}{%
  \begin{tabular}{l ccc ccc ccc ccc}
    \toprule
    \multirow{2}{*}{Method} & \multicolumn{3}{c}{1-way 1-shot} & \multicolumn{3}{c}{1-way 5-shot} & \multicolumn{3}{c}{2-way 1-shot} & \multicolumn{3}{c}{2-way 5-shot} \\
    \cmidrule(lr){2-4} \cmidrule(lr){5-7} \cmidrule(lr){8-10} \cmidrule(lr){11-13}
    & $S^0$ & $S^1$ & Mean & $S^0$ & $S^1$ & Mean & $S^0$ & $S^1$ & Mean & $S^0$ & $S^1$ & Mean \\
    \midrule
    \multicolumn{13}{l}{\textit{Official Results (from Literature)}} \\
    AttMPTI~\cite{zhao2021few} & 34.03 & 30.97 & 32.50 & 39.09 & 37.15 & 38.12 & 25.99 & 23.88 & 24.94 & 30.41 & 27.35 & 28.88 \\
    QGE~\cite{ning2023boosting} & 37.38 & 33.02 & 35.20 & 45.08 & 41.89 & 43.49 & 26.85 & 25.17 & 26.01 & 28.35 & 31.49 & 29.92 \\
    COSeg~\cite{an2024rethinking} & 41.95 & 42.07 & 42.01 & 48.54 & 44.68 & 46.61 & 29.54 & 28.51 & 29.03 & 36.87 & 34.15 & 35.51 \\
    \textcolor{gray}{MM-FSS~\cite{an2024multimodality}} & \textcolor{gray}{46.08} & \textcolor{gray}{43.37} & \textcolor{gray}{44.73} & \textcolor{gray}{54.66} & \textcolor{gray}{45.48} & \textcolor{gray}{50.07} & \textcolor{gray}{43.99} & \textcolor{gray}{34.43} & \textcolor{gray}{39.21} & \textcolor{gray}{48.86} & \textcolor{gray}{39.32} & \textcolor{gray}{44.09} \\
    \midrule
    \multicolumn{13}{l}{\textit{Reproduced Environment (Strict Control)}} \\
    MM-FSS$^{\dagger}$ (Official Weights) & 45.68 & \textbf{43.23} & 44.46 & 54.80 & 45.79 & 50.30 & 44.01 & 34.44 & 39.23 & 49.11 & 39.48 & 44.20 \\ 
    \textbf{Ours (DA-FSS)} 
    & \textbf{48.95} & 41.96 & \textbf{45.46}$_{\textcolor{DarkGreen}{\uparrow\textbf{1.00}}}$
    & \textbf{55.40} & \textbf{46.30} & \textbf{50.85}$_{\textcolor{DarkGreen}{\uparrow\textbf{0.55}}}$ 
    & \textbf{45.01} & \textbf{34.53} & \textbf{39.77}$_{\textcolor{DarkGreen}{\uparrow\textbf{0.54}}}$
    & \textbf{50.61} & \textbf{40.45} & \textbf{45.53}$_{\textcolor{DarkGreen}{\uparrow\textbf{1.33}}}$ \\
    \bottomrule
  \end{tabular}
  }
  \caption{\textbf{Quantitative results on ScanNet.} \textit{Top:} Literature results. \textit{Bottom:} Comparison under our strictly controlled evaluation protocol using official pre-trained weights. mIoU results are reported in \%. The best performance in the controlled group is highlighted in \textbf{bold}.}
  \label{tab:sota_scannet}
\end{table*}

\begin{figure*}[t]
  \centering
  \includegraphics[width=1.0\linewidth]{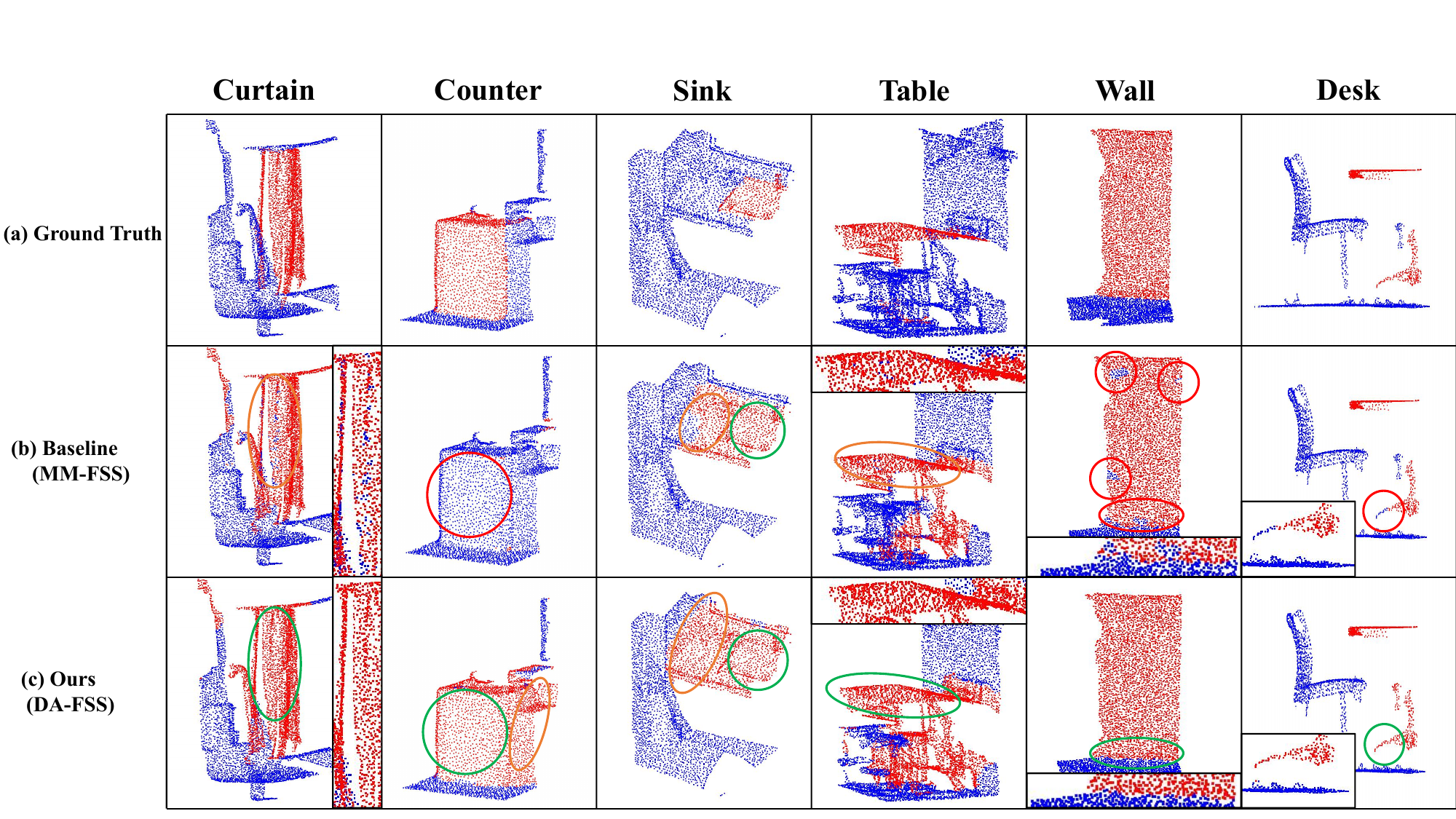} 
        \caption{Qualitative comparison between MM-FSS and our proposed DA-FSS in the 1-way 1-shot setting on the ScanNet dataset split 1. The target classes in the columns are curtain, counter, sink, table, wall and desk, respectively. Comparison between (b) Baseline and (c) Ours demonstrates a fundamental trade-off. It is evident that Ours (DA-FSS) exhibits a significant advantage in the geometric completeness of classes and can effectively repair instances of missing points. Moreover, on counter classes, Baseline (MM-FSS) chooses to directly ignore them, whereas Ours (DA-FSS) is capable of recognizing them. However, Ours occasionally exhibits boundary over-segmentation to ensure geometric completeness.}
  \label{fig:tradeoff}
\end{figure*}

\subsection{Stacked Arbitration Module (SAM)}
\label{subsec:arbitration}
While the two experts can make predictions independently; without fusing and refining these two predictions, the multimodal system cannot fully realize its potential for cross-checking and compensating for deficiencies. To synthesize the final prediction from the refined experts, we propose the Stacked Arbitration Module (SAM). Unlike prior works that rely on simple concatenation, we opt for a \textbf{Stacked Transformer Arbitration} architecture to model global interactions between geometric and semantic evidences.

\paragraph{Feature Merging Layer.}
We first project the concatenated expert outputs into a unified space (192-dimensional). 
Consistent with our implementation, Batch Normalization (BN) is applied pre-fusion to calibrate the scale discrepancy between adaptive and static features, followed by a $1 \times 1$ convolution and ReLU activation:
\begin{equation}
    R_{merged} = \text{ReLU}(\mathcal{F}_{conv}(\text{BN}([R_{geo} \parallel R_{sem}])))
\end{equation}

\paragraph{Dual-Guidance Injection.}
We inject inductive biases strictly at the boundaries using a "Negative-Suppression, Positive-Enhancement" strategy:
\begin{itemize}[leftmargin=*]
    \item \textbf{Input: Background Suppression ($G_{base}$).} 
    At the input of every Transformer layer, we explicitly inject Base-Class Guidance ($G_{base}$) to refine the background representation. 
    Specifically, we concatenate $G_{base}$ with the background token $R_{bg}$ to suppress easy negatives early, while leaving foreground tokens $R_{fg}$ untouched:
    \begin{equation}
        R_{bg}^{(1)} = \mathcal{F}_{lin}([R_{bg} \parallel G_{base}]), \quad R_{in}^{(1)} = [R_{bg}^{(1)} \parallel R_{fg}]
    \end{equation}
This can lend the model a more hierarchical structure, enabling further refinement of the feature space.
    
    \item \textbf{Output: Semantic Gating ($G_q$).} 
    At the final layer, we employ a multiplicative gating mechanism derived from the target class embedding ($G_q$). 
    This acts as a semantic gate to enhance foreground precision by scaling the feature magnitude based on textual affinity:
    \begin{equation}
        R_{final} = R_{arb}^{(N)} \odot (1 + \sigma(G_q))
    \end{equation}
    where $\odot$ denotes element-wise multiplication and $\sigma$ is a scaling function.
\end{itemize}
This effectively mitigates issues of over-suppression or over-amplification while also reducing the parameter overhead in deep stacks beyond two layers.

\subsection{Optimization Objective} 
\label{subsec:optimization} 

After the Stacked Arbitration Module (SAM), the synthesized feature tensor $R_{final}$ is transformed into the final prediction $\hat{P}_q \in \mathbb{R}^{N_q \times (N_{way}+1)}$ by a decoder comprising a KPConv~\cite{thomas2019kpconv} layer and an MLP classifier. The whole model is optimized end-to-end by minimizing a compound objective function. To ensure expert coordination without compromising individual plasticity, we integrate the segmentation task with decoupled alignment constraints: 


\begin{equation}
\begin{split}
\mathcal{L}_{total} = & \mathcal{L}_{seg} + \lambda_{base}\mathcal{L}_{base} \\
& + \lambda_{PLR}\mathcal{L}_{PLR} + \lambda_{DCR}\mathcal{L}_{DCR},
\end{split}
\label{eq:total_loss}
\end{equation}

where $\mathcal{L}_{seg}$ is the standard cross-entropy loss between $\hat{P}_q$ and the ground-truth $Y_q$, and $\mathcal{L}_{base}$ checks the predictions for the base class.  This formulation ensures a cooperative optimization: vertical gradients from $\mathcal{L}_{seg}$ optimize the decision boundary, while horizontal gradients from the alignment terms maintain expert coordination.

\section{Experiments}
\label{sec:experiments}

In this section, we conduct extensive experiments to validate our proposed Decoupled-experts Arbitration (DA-FSS) framework. We first introduce the experimental setup.
Then, we compare our method with state-of-the-art (SOTA) approaches on two challenging benchmarks: S3DIS~\cite{armeni20163d} and ScanNet~\cite{dai2017scannet}.
Finally, we perform detailed ablation studies to demonstrate the effectiveness of each key component in our architecture.

\subsection{Experimental Setup}
\paragraph{Datasets.}
We evaluate our method on two standard few-shot 3D point cloud segmentation benchmarks: S3DIS~\cite{armeni20163d} and ScanNet~\cite{dai2017scannet}.
\textbf{S3DIS} contains 3D scans from 6 large-scale indoor areas. 
\textbf{ScanNet} consists of 1513 scanned indoor scenes. We adopt the official benchmark splits~\cite{dai2017scannet}, which contain 20 semantic classes.
Their semantic classes are split into 2 folds ($S^0$ and $S^1$) for cross-validation.
Following prior works~\cite{an2024multimodality,zhao2021few}, we maintain a maximum of 20,480 points per block using a 0.02m grid size.

\paragraph{Evaluation Protocol.}
We follow the standard $N$-way $K$-shot episodic evaluation protocol.
The primary evaluation metric is the mean Intersection-over-Union (mIoU) over all foreground classes.
As in ~\cite{an2024multimodality,an2024rethinking}, we report results on 1-way 1-shot, 1-way 5-shot, 2-way 1-shot, and 2-way 5-shot settings where the evaluation sets consist of 1,000 episodes per class in the 1-way setting and 100 episodes per class combination in the 2-way setting.

\paragraph{Implementation Details.}
We adopt the same backbone (Stratified Transformer) ~\cite{lai2022stratified} and feature head architecture as MM-FSS~\cite{an2024multimodality} for fair comparison.
Following the standard protocol~\cite{an2024multimodality}, we initialize the 3D backbone with pre-trained weights that use the AdamW optimizer, setting a weight decay of 0.01 and a learning rate of 0.006 during pretraining in 100 epochs. Isolating the source of improvement, we choose to download officially released pre-trained weights directly. During the meta-learning phase, the model is trained using the AdamW optimizer with an initial learning rate of $0.0001$ and a weight decay of $0.01$.
For our DA-FSS architecture, we set the number of layers in the Stacked Arbitration Module (SAM) to $N=1$ for S3DIS and $N=2$ for ScanNet.
For the Decoupled Alignment Module (DAM), the loss weights are set to $\lambda_{PLR} = 0.001$ and $\lambda_{DCR} = 0.5$ based on empirical tuning.

\subsection{Comparison with State-of-the-Art}
\label{sec:comparison}

We compare our DA-FSS with previous state-of-the-art methods on S3DIS~\cite{armeni20163d} and ScanNet~\cite{dai2017scannet}, as detailed in Table~\ref{tab:sota_s3dis} and Table~\ref{tab:sota_scannet}.
To ensure a fair assessment, we reproduced the leading baseline MM-FSS~\cite{an2024multimodality} (denoted as MM-FSS$^{\dagger}$) using its officially released pre-trained weights under our strictly controlled evaluation protocol.
In contrast, DA-FSS consistently outperforms the former state-of-the-art across all settings, demonstrating the improved stable generalization brought by decoupling, which enhances the utilization rate of information from pre-trained weights.

Specifically, on the S3DIS dataset, DA-FSS records mIoU increases of \textbf{+1.21\%} in the 1-way-1-shot setting—which relies heavily on intrinsic geometric generalization. Furthermore, in 1-way-1-shot setting on the ScanNet dataset, DA-FSS has also achieved \textbf{+1.00\%} improvement over MM-FSS. Notably, DA-FSS surpasses the baseline by \textbf{+1.33\%} in the 2-way 5-shot setting (see Supp. Material for full tables) that our Decoupled Arbitration mechanism effectively disentangles inter-class semantics under high interference, while simultaneously enforcing intra-class prototype stability through alignment constraints.

Remarkably, beyond standard mIoU metrics, our DA-FSS demonstrates a substantial improvement in \textbf{Mean Accuracy (mAcc)} on ScanNet. 
In settings where mIoU gains appear moderate (e.g., ScanNet 1-way 1-shot), our mAcc often surges by approximately \textbf{+10\%} compared to the baseline shown in Table~\ref{tab:main_ablation}(c).

Although in some cases ours (DA-FSS) may induce over-segmentation due to its tendency to preserve geometric completeness with high-variance class confusion (which penalizes IoU heavily), this very tendency actually aids the model in more accurately segmenting points that should have been correctly identified. We prioritize semantic completeness over boundary precision, arguing that successfully identifying the target outweighs the cost of minor over-segmentation in 1-way 1-shot setting.

\begin{table}[t]
    \centering

    \renewcommand{\arraystretch}{1.3} 
    \setlength{\tabcolsep}{2pt}

    \resizebox{\linewidth}{!}{%
        \begin{tabular}{llcc}
            \toprule
            \textbf{Config.} & \textbf{Architecture Detail} & \textbf{DAM} & \textbf{mIoU (\%)} \\
            \midrule
            (A) Base & MM-FSS (Fused)~\cite{an2024multimodality} & \ding{55} & 44.46 \\
            (B) Ours & Core Arch. (Decoupled) & \ding{55} & 44.95 \\
            (C) \textbf{Ours} & \textbf{Full Model (Decoupled+DAM)} & \ding{51} & \textbf{45.46} \\
            \bottomrule
        \end{tabular}
    }
    \vspace{1pt} 
    \centerline{\footnotesize (a) Architectural Effectiveness}

    \vspace{6pt}

    \resizebox{\linewidth}{!}{%
        \begin{tabular}{lccc}
            \toprule
            \textbf{Method} & \textbf{FLOPs (G)} & \textbf{Params (M)} & \textbf{mIoU (\%)} \\
            \midrule
            MM-FSS$^{\dagger}$ & 19.06 & 10.45 & 44.46 \\
            \textbf{Ours (DA-FSS)} & \textbf{18.76} & \textbf{10.18} & \textbf{45.46} \\
            \midrule
            \textit{Diff.} & 
            \textcolor{DarkGreen}{\textbf{-0.30}} & 
            \textcolor{DarkGreen}{\textbf{-0.27}} & 
            \textcolor{DarkGreen}{\textbf{+1.00}} \\
            \bottomrule
        \end{tabular}
    }
    \vspace{1pt}
    \centerline{\footnotesize (b) Complexity Analysis}

    \vspace{6pt}

    \resizebox{\linewidth}{!}{
        \begin{tabular}{lccl} 
            \toprule
            \textbf{Method} & \textbf{mIoU} & \textbf{mAcc(\%)} & \textbf{Mechanism Explanation} \\
            \midrule
            MM-FSS$^{\dagger}$ & 44.46 & 68.33$^*$ & \textit{Catastrophic Miss (High FN)} \\
            \textbf{Ours} & \textbf{45.46} & \textbf{79.29}$^*$ & \textbf{Object Recovered (High Recall)} \\
            \midrule
            \textit{Gain} & \textcolor{DarkGreen}{\textbf{+1.0}} & \textcolor{DarkGreen}{\textbf{+10.9}} & \textbf{\textit{Solves Semantic Blindness}} \\
            \bottomrule
        \end{tabular}
    }
    \vspace{1pt}
    \centerline{\footnotesize (c) \textbf{The Decoupling Effect}}
    \caption{Comprehensive analysis on ScanNet (1-way 1-shot). We analyze (a) architectural effectiveness, (b) computational complexity, and (c) the specific impact of decoupling.
    \label{tab:main_ablation}
    \textbf{Note:} DA-FSS achieves a massive surge in mAcc.}
\end{table}

\subsection{Ablation Studies} 
\label{sec:ablation} 
In this section, we conduct a series of ablation studies. Unless specified otherwise, all ablation experiments are performed on the \textbf{ScanNet} dataset under the \textbf{1-way 1-shot} setting, as this setting is most sensitive to feature quality and generalization capability.

\paragraph{Effectiveness of DA-FSS Architecture.} Table~\ref{tab:main_ablation}(a) validates our core hypothesis: "Decoupling" is structurally superior to "Fusion." Comparing (A) and (B), we observe that mIoU increases by approximately \textbf{+0.49\%} solely through the decoupled design, even when all other components are disabled. This confirms that simply removing the early fusion bottleneck brings gains by mitigating the plasticity-stability conflict. Comparing (B) and (C), we achieved a \textbf{+0.51\%} improvement after enabling all modules. This proves that while decoupling is essential, the experts must be coordinated via regularization (our DAM) to prevent divergence and achieve optimal alignment thereby optimizing the utilization of information interaction across various modalities..

\paragraph{Complexity Analysis.} Finally, Table~\ref{tab:main_ablation}(b) highlights the efficiency of our design. Our DA-FSS achieves a significant performance gain (\textbf{+1.00\%} mIoU) over the MM-FSS baseline while actually reducing computational overhead (\textbf{-0.30} GFLOPs) and parameters (\textbf{-0.27} M). This confirms that our gains stem from the superior decoupled architecture, not from increased model capacity.

\section{Conclusion}
In this paper, we identify a critical conflict in FS-PCS: the "Plasticity-Stability Dilemma," which has undermined the validity of previous progress in fusion-based paradigms. To rectify this issue, we present DA-FSS designed to decouple geometric adaptation from semantic preservation to maximize its adaptability across modalities. DA-FSS combines Parallel Experts to effectively aggregate features, enriching the comprehension of novel concepts from both geometric and semantic perspectives, which are mutually important for recovering missing parts. Furthermore, to ensure expert coordination despite decoupling, we introduce the DAM, which mutually regularizes gradients via prototype alignment. Finally, we introduce the SAM to fuse dual-path features and dynamically synthesize final decisions using boundary-injected guidance. DA-FSS achieves significant improvements over existing methods across all settings, particularly in boosting point-wise accuracy (mAcc). Overall, our research provides valuable insights into the importance of Decoupled Arbitration in FS-PCS and suggests promising directions for future studies.

{
    \small
    \bibliographystyle{named}
    \bibliography{ijcai26}
}
\end{document}